\documentclass[runningheads]{llncs}

\usepackage{eccv}

\usepackage{eccvabbrv}

\usepackage{graphicx}
\usepackage{booktabs}

\usepackage[accsupp]{axessibility}  

\usepackage{adjustbox} 
\usepackage{multirow}
\usepackage{array} 
\usepackage[table]{xcolor}

\usepackage{hyperref}

\usepackage{orcidlink}

\begin{document}

\title{\textit{EraseSAE}: Surgical Concept Erasure in Text-to-Video Diffusion Models via Sparse Autoencoders}  

\titlerunning{\textit{EraseSAE}}

\author{Xinghao Wang\inst{1,3}$^\ast$\orcidlink{0009-0002-2675-6781} \and
Dong Li\inst{2}$^\dagger$\orcidlink{0000-0002-9936-1126} \and
Wei Yu\inst{2}\orcidlink{0000-0001-6478-3903} \and
Yingwei Pan\inst{2}\orcidlink{0000-0002-4344-8898} \and
Tao Gong\inst{1,3}$^\dagger$\orcidlink{0000-0003-0026-6813} \and
Qi Chu\inst{1,3}\orcidlink{0000-0003-3028-0755} \and
Nenghai Yu\inst{1,3}\orcidlink{0000-0003-4417-9316} \and
Ting Yao\inst{2}\orcidlink{0000-0001-7587-101X} 
}

\authorrunning{X.~Wang et al.}

\institute{University of Science and Technology of China, Hefei, China \and
HiDream.ai Inc. \and
Anhui Province Key Laboratory of Digital Security, China\\
\email{wxhwxhwxh@mail.ustc.edu.cn, \{lidong,yuwei,pandy\}@hidream.ai, \{tgong,qchu,ynh\}@ustc.edu.cn, tiyao@hidream.ai}}

\maketitle

\footnotetext[1]{This work was performed at HiDream.ai. $^\dagger$ Corresponding authors.}
\begin{abstract}
    Recent advances in text-to-video (T2V) diffusion models have demonstrated remarkable generative capabilities, yet their reliance on loosely curated training data raises pressing safety and copyright concerns. Concept erasure offers a principled remedy by removing unwanted semantics from pretrained models while preserving remaining concepts. However, existing approaches typically operate at a coarse granularity misaligned with the fine-grained, distributed nature of concept representations, leading to incomplete removal or degraded generation quality. We argue that surgical erasure fundamentally requires intervention at the level of monosemantic features, where each unit encodes a single interpretable concept. To this end, we propose \textit{EraseSAE}, a novel framework that leverages sparse autoencoders to achieve surgical concept erasure in DiT-based T2V diffusion models via a principled \textit{decompose-attribute-erase} pipeline. We first introduce the Partitioned Convolutional Sparse Autoencoder, which decomposes dense spatiotemporal activations into disentangled, interpretable sparse features while preserving spatiotemporal coherence. A contrastive attribution mechanism then contrasts activations from paired prompts to isolate concept-specific feature kernels. At inference, timestep-resolved spatiotemporal masks derived from the identified kernels confine erasure to regions where the target concept is active, leaving unrelated content intact. Extensive experiments across diverse diffusion models and concept erasure tasks demonstrate that \textit{EraseSAE} achieves precise and robust concept removal with minimal quality degradation, substantially outperforming state-of-the-art methods. The code is available at \url{https://github.com/HiDream-ai/EraseSAE}.
\end{abstract}

\section{Introduction}
\label{sec:intro}
The rapid advancement of T2V diffusion models~\cite{yang2025cogvideox,kong2024hunyuanvideo} has fundamentally transformed video creation, enabling high-fidelity video synthesis from free-form text prompts. However, this powerful generative capability also introduces significant safety and copyright risks~\cite{schramowski2023safe,jiang2023ai}. Trained on massive and loosely curated datasets, these models can readily generate harmful content such as explicit material or deepfakes of public figures. Mitigating such unsafe outputs has emerged as a critical challenge for responsible deployment of T2V models. While retraining on filtered datasets is a straightforward solution~\cite{huang2025survey}, the prohibitive computational cost renders it impractical. Concept erasure~\cite{xu2025videoeraser,yoon2025safree,lu2024mace} provides a viable alternative by selectively removing target semantics from a pretrained model while preserving its capacity to generate remaining content. However, as illustrated in Fig.~\ref{fig:q_nud}, achieving surgical erasure that precisely removes target concepts without degrading broader generative capabilities remains an open problem.

Existing concept erasure methods, primarily adapted from the text-to-image (T2I) domain, generally fall into two paradigms: \textit{training-free} and \textit{training-based}. \textit{Training-free} methods steer inference away from target concepts through techniques such as negative prompting~\cite{rombach2022high}, text embedding manipulation~\cite{yoon2025safree,xu2025videoeraser,li2024get}, or conditional guidance~\cite{schramowski2023safe,zhang2025minimalist}, without altering model weights. While computationally efficient, they merely suppress concepts at the surface level while leaving internal concept representations intact, rendering them vulnerable to adversarial attacks. \textit{Training-based} methods seek stronger guarantees by permanently altering model parameters through global fine-tuning~\cite{ye2025t2vunlearning,gao2025eraseanything,gandikota2023erasing,kumari2023ablating}, closed-form editing~\cite{lu2024mace}, or neuron pruning~\cite{yang2024pruning}. Although more robust to input manipulation, these methods suffer from a fundamental granularity mismatch: existing methods operate at a coarse granularity that fails to align with the fine-grained, distributed nature of concept representations within these models. In deep neural networks, concepts are distributed across \textit{polysemantic} representations where individual neurons encode multiple concepts simultaneously. Intervening on such entangled units leads to three distinct failure modes: (i) global weight modifications cause catastrophic forgetting of unrelated concepts; (ii) layer-specific edits targeting only cross-attention projections leave concept traces in other pathways, enabling reconstruction under adversarial probing; and (iii) neuron-level pruning inflicts collateral damage on co-encoded benign concepts. Beyond this granularity mismatch, virtually all existing methods lack spatiotemporal locality: even when a target concept occupies only a local spatial region or temporal segment, current approaches apply erasure globally, unnecessarily degrading unrelated content (Fig.~\ref{fig:q_nud}). This limitation is particularly acute in video generation, where 3D full-attention in DiT-based T2V models encodes concepts jointly across space and time, making them especially sensitive to coarse-grained intervention.

These limitations converge on a fundamental requirement: surgical concept erasure demands intervention at the level of \textit{monosemantic} units, where each feature encodes a single, interpretable concept. Sparse autoencoders (SAEs)~\cite{olshausen1997sparse}, recently developed for mechanistic interpretability of large language models (LLMs)~\cite{cunningham2024sparse,shi2025route}, provide exactly such a substrate by decomposing high-dimensional activations into sparse linear combinations of monosemantic features. This property offers three key advantages that directly address the above limitations. First, projecting dense activations into a disentangled sparse space enables isolating specific features without disturbing others, resolving the polysemanticity problem that plagues neuron-level pruning. Second, SAE features capture distributed representations spanning arbitrary network components, overcoming the incomplete coverage of layer-specific editing. Third, because SAE activations naturally vary across spatial positions and frames, suppressing a target feature automatically confines erasure to where the concept is active, thereby achieving spatiotemporal locality that global fine-tuning cannot provide. Despite these advantages, transferring SAEs from LLMs to DiT-based T2V models presents substantial challenges. Conventional linear SAEs neglect local dependencies across neighboring positions and frames, risking degradation of spatiotemporal coherence. Furthermore, whereas LLM activations maintain stable semantics once generated, diffusion features progressively transition from coarse structure to fine detail across denoising timesteps, rendering single-timestep attribution insufficient.

Motivated by these insights, we propose \textit{EraseSAE}, a novel framework that leverages SAEs to achieve surgical concept erasure in DiT-based T2V diffusion models via a principled \emph{decompose--attribute--erase} pipeline. In the \textit{decompose} stage, we introduce the \textit{Partitioned Convolutional Sparse Autoencoder (PConvSAE)} to address the challenge of spatiotemporal coupling. By employing multi-dimensional convolutions and a partitioned architecture, PConvSAE decomposes coupled spatiotemporal activations into interpretable, concept-specific sparse features while preserving spatial structure and temporal coherence. In the \textit{attribute} stage, we feed paired prompts that include and exclude the target concept into the frozen diffusion model and contrast the resulting activation distributions within PConvSAE, isolating a compact set of feature kernels highly correlated with the target concept. In the \textit{erase} stage, we derive spatiotemporal masks from the activation maps of the identified kernels at each denoising timestep, and selectively suppress the target concept by modifying PConvSAE features only within the masked regions. This strategy restricts interventions exclusively to regions where the target concept is active while preserving the original feature evolution elsewhere. By operating in the monosemantic feature space, \textit{EraseSAE} unifies the strengths of both paradigms: it achieves the robust and permanent erasure of training-based methods while maintaining the surgical precision and minimal side effects that training-free methods aspire to but cannot guarantee. More broadly, our work establishes that mechanistic interpretability tools can serve as a foundation for precise and controllable generation in the video domain.

We summarize our main contributions as follows:
\begin{itemize}
    \item We propose \textit{EraseSAE}, the first framework to leverage SAEs for concept erasure in T2V diffusion models, which surgically removes target concepts through a \emph{decompose--attribute--erase} pipeline in monosemantic feature space.
    \item We introduce PConvSAE, a convolutional sparse autoencoder with a partitioned architecture that decomposes dense visual representations into interpretable sparse features while preserving spatiotemporal coherence.
    \item We design a contrastive attribution mechanism that isolates concept-correlated features and derives timestep-resolved spatiotemporal masks, confining erasure to where target concepts are active while preserving non-target content.
    \item Extensive experiments across multiple T2V diffusion models and diverse concept erasure tasks demonstrate that \textit{EraseSAE} achieves state-of-the-art erasure effectiveness while preserving generation quality, outperforming the strongest baseline by 34.5\% in erasure accuracy and 8.3\% in SSIM.
\end{itemize}

\section{Related Works}

\subsection{Concept Erasure in Diffusion Models}
Concept erasure has been extensively studied in T2I diffusion models~\cite{rombach2022high,flux2024,cai2025hidream,cai2026hidream} along two main paradigms. \textit{Training-free} methods~\cite{zhang2025minimalist,lee2025localized,na2025training,li2024get} redirect inference without modifying model weights. SLD~\cite{schramowski2023safe} introduces a safety guidance term to steer denoising away from unsafe content, while SAFREE~\cite{yoon2025safree} projects text embeddings orthogonally to toxic concept directions. Though efficient, these approaches suppress concepts only at the surface level, leaving internal representations vulnerable to adversarial attacks~\cite{hsu2024ring,chin2024prompting4debugging}. \textit{Training-based} methods offer stronger guarantees by permanently altering model parameters. ESD~\cite{gandikota2023erasing} fine-tunes cross-attention layers with negative-prompt guidance, and CA~\cite{kumari2023ablating} ablates target concepts by anchoring outputs to a reference distribution. UCE~\cite{gandikota2024unified} unifies multi-concept editing via closed-form projection updates, while AdvUnlearn~\cite{zhang2024defensive} and Receler~\cite{huang2024receler} augment fine-tuning with adversarial training for improved robustness. MACE~\cite{lu2024mace} further scales erasure to massive concept sets through closed-form parameter mapping, and EraseAnything~\cite{gao2025eraseanything} extends the paradigm to flow-matching architectures via bi-level optimization. However, these methods operate at a coarse granularity and apply erasure globally, lacking the locality needed to preserve unrelated content. Extension to T2V models remains nascent. VideoEraser~\cite{xu2025videoeraser} projects prompt embeddings away from unsafe directions, and T2VUnlearning~\cite{ye2025t2vunlearning} fine-tunes velocity predictions with negative guidance. However, the spatiotemporal entanglement inherent in video generation amplifies the inherited limitations from T2I methods, and no existing approach achieves fine-grained, localized erasure without degrading overall generation quality.

\subsection{Sparse Autoencoders for Mechanistic Interpretability}
SAEs decompose dense neural activations into sparse linear combinations of monosemantic features, providing a principled tool for mechanistic interpretability~\cite{olshausen1997sparse}. Originally developed for modeling biological vision, SAEs have recently shown notable success in interpreting large-scale transformer-based language models~\cite{cunningham2024sparse}, with subsequent work scaling to frontier models~\cite{templeton2024scaling} and enabling feature-level steering for controllable generation~\cite{shi2025route}. Recent efforts have extended SAEs to visual generative models. SAUCE~\cite{geng2025sauce} applies SAEs within autoregressive vision-language models to suppress undesirable features for selective concept unlearning. SAeUron~\cite{cywinski2025saeuron} integrates SAEs into image diffusion models, achieving interpretable concept erasure through feature-level interventions. However, no prior work has investigated SAEs in T2V diffusion models. Conventional MLP-based SAE architectures flatten high-dimensional hidden states into one-dimensional vectors, compromising the essential spatiotemporal locality inherent in video representations. Furthermore, standard SAEs lack mechanisms for rigorous semantic isolation across spatial regions, leading to concept leakage between target and non-target content. Our proposed PConvSAE addresses both limitations through a convolutional architecture that preserves full spatiotemporal structure and a partitioned design for high-purity feature disentanglement.

\section{Methodology}
\begin{figure}[t]
\centering
\captionsetup{font=small}
\includegraphics[width=0.98\textwidth]{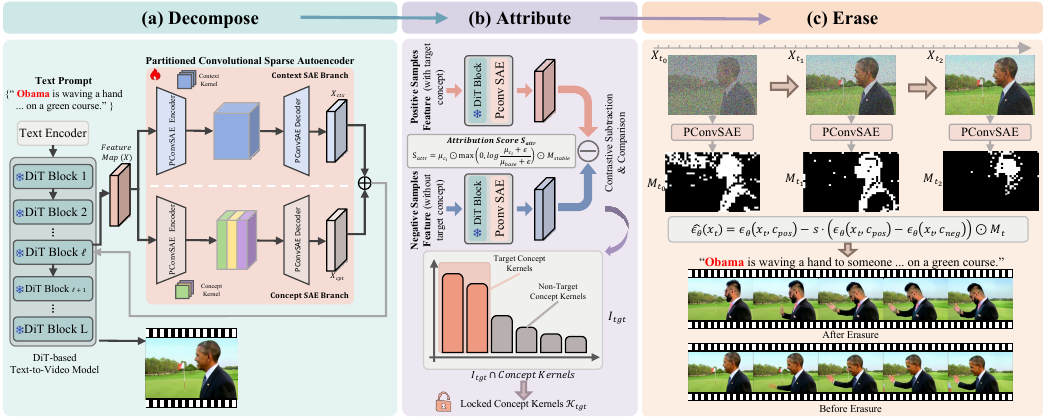}
\caption{Overview of \textit{EraseSAE} framework. \textit{EraseSAE} follows a \emph{decompose--attribute--erase} pipeline. (a)~\textit{Decompose}: PConvSAE decomposes dense spatiotemporal activations into disentangled monosemantic features via a partitioned dual-branch convolutional architecture with spatial-aware activation. (b)~\textit{Attribute}: a contrastive log-ratio scoring mechanism isolates concept-specific feature kernels against hard-negative baselines in a one-time offline procedure. (c)~\textit{Erase}: timestep-resolved spatiotemporal masks derived from the locked kernels guide spatially-modulated classifier-free guidance, confining suppression to active concept regions while preserving unrelated content intact.}
\label{fig:pipline}
\end{figure}

\subsection{Overview of \textit{EraseSAE}}
Given a pretrained DiT-based T2V diffusion model and a set of target concepts, \textit{EraseSAE} surgically removes specified semantics from generated videos while preserving unrelated content. The framework follows a \emph{decompose--attribute--erase} pipeline, as illustrated in Fig.~\ref{fig:pipline}. In the \textit{decompose} stage (Sec.~\ref{sec:pconvsae}), we train a PConvSAE on intermediate activations at an identified intervention layer. Through multi-dimensional convolutions and a strictly partitioned dual-branch architecture, PConvSAE transforms dense spatiotemporal representations into disentangled monosemantic features, separating concept-specific semantics from general scene context. In the \textit{attribute} stage (Sec.~\ref{sec:attr}), a one-time contrastive procedure contrasts activation distributions from paired prompts that include and exclude the target concept. Log-ratio scoring against hard-negative baselines isolates a compact set of concept-specific feature kernels with high semantic purity. In the \textit{erase} stage (Sec.~\ref{sec:erase}), the identified kernels produce timestep-resolved spatiotemporal masks that dynamically track the target concept throughout the denoising process. A spatially-modulated classifier-free guidance mechanism uses these masks to confine suppression exclusively to active concept regions.

\subsection{Partitioned Convolutional Sparse Autoencoder}
\label{sec:pconvsae}

In DiT-based video diffusion models, intermediate hidden states encode densely entangled spatiotemporal representations where semantic content is distributed across spatial positions and temporal frames in a tightly coupled manner. Conventional MLP-based SAEs~\cite{cywinski2025saeuron} flatten these high-dimensional hidden states into one-dimensional vectors prior to sparse decomposition, destroying the inherent spatial structure and temporal continuity of video features. This degrades reconstruction fidelity and impairs concept localization in both space and time. 

To address these challenges, we first conduct hierarchical feature probing across transformer layers to identify the optimal intervention layer that captures rich target semantics while minimally affecting global structural fidelity. At this identified layer, we introduce PConvSAE, which replaces the standard linear projection with a stack of 2D convolutions operating on temporally folded activation tensors. By folding the temporal dimension into the batch axis and applying spatial convolutions directly on the resulting tensor, PConvSAE encodes and reconstructs features on their native topology, preserving the complete spatiotemporal structure without the information loss inherent in flattening operations.

\textbf{Architecture Design.} The central design principle of PConvSAE is the explicit decomposition of entangled visual representations into two complementary and functionally distinct components: concept-agnostic scene context and concept-specific semantics. Accordingly, PConvSAE partitions its latent space into two structurally decoupled computational branches: a \textit{Context Branch} for encoding concept-agnostic scene representations $f_{\text{ctx}}$, and a \textit{Concept Branch} for isolating concept-specific features $f_{\text{cpt}}$. The Context Branch captures global structural priors, background layout, illumination, and temporal dynamics that constitute the scene canvas, while the Concept Branch exclusively accommodates localized features semantically bound to target concepts. Together, they form a complete and non-redundant decomposition: suppressing Concept Branch features removes only target semantics while the Context Branch preserves all remaining visual content. Within the Concept Branch, we further enforce concept partitioning by dividing the channel dimension into $C$ mutually exclusive subspaces (one per target concept, each with $N_{\text{cpt}}$ feature kernels), preventing semantic interference during attribution and erasure.

Given the debiased hidden state $X \in \mathbb{R}^{(B \times T) \times D \times H \times W}$ obtained by subtracting the running mean of activations, both branches project $X$ into a high-dimensional latent space via independent spatial convolutional encoders:
\begin{equation}
z_{\text{ctx}} = W_{\text{enc}}^{\text{ctx}} * X + b_{\text{enc}}^{\text{ctx}}, \quad z_{\text{cpt}} = W_{\text{enc}}^{\text{cpt}} * X + b_{\text{enc}}^{\text{cpt}}
\label{eq:0}
\end{equation}
where $W_{\text{enc}}$ and $b_{\text{enc}}$ denote the convolutional weights and biases of the respective encoders, and $*$ represents the convolution operation. The two encoders do not share parameters, ensuring entirely independent projection subspaces.

To impose sparsity while respecting the spatial nature of visual concepts, we introduce a Spatial-Aware Local Activation mechanism. Rather than applying a conventional channel-wise Top-K operator, we evaluate each channel $c$ by its peak spatial response $s_c = \max_{h,w} z_{c,h,w}$ and select the index set $\mathcal{I}_{\text{top}}$ of the $K$ most responsive channels. The sparse feature representation is then defined as:
\begin{equation}
f_{c, h, w} = \begin{cases} \text{ReLU}(z_{c, h, w}), & \text{if } c \in \mathcal{I}_{\text{top}} \\ 0, & \text{otherwise} \end{cases}
\label{eq:1}
\end{equation}
where $c$, $h$, and $w$ index the channel, height, and width, respectively. This formulation activates only the most salient channels while suppressing others across their entire spatial extent, producing spatially coherent sparse codes. The mechanism is applied independently to both branches with separate sparsity budgets, allowing the Context Branch to retain broader scene-level features while the Concept Branch maintains highly selective concept-correlated activations.

The activated features are projected back via respective convolutional decoders, and the overall reconstruction is obtained by additive superposition:
\begin{equation}
X_{\text{recon}} = W_{\text{dec}}^{\text{ctx}} * f_{\text{ctx}} + W_{\text{dec}}^{\text{cpt}} * f_{\text{cpt}} + b_{\text{dec}}
\end{equation}
This additive formulation embodies the complementary principle: at erasure time, removing the Concept Branch contribution while retaining the Context Branch naturally yields a clean scene with the target concept excised.

\textbf{Joint Optimization.} Training PConvSAE requires guiding the two branches toward their designated complementary roles. We formulate a joint optimization objective supervised by spatial masks derived from the diffusion model's own cross-attention maps, eliminating the need for external annotations. Specifically, we extract cross-attention maps between visual hidden states and textual embeddings of the target concept~\cite{cai2025ditctrl, gong2026freeinpaint}, average and normalize them across selected layers, and apply dynamic quantile-based thresholding to yield a binarized target region mask $M_{\text{tgt}} \in \{0, 1\}$,  with the background mask defined as its complement $M_{\text{bg}} = 1 - M_{\text{tgt}}$. Guided by this pair of mutually exclusive spatial priors, the overall optimization objective comprises five components.

The Context Reconstruction Loss ($\mathcal{L}_{\text{ctx}}$) compels the Context Branch to faithfully reconstruct the background region while being suppressed within the target spatial extent, steering it toward encoding only concept-agnostic content:
\begin{equation}
\mathcal{L}_{\text{ctx}} = \lambda_{\text{bg}} \frac{\| M_{\text{bg}} \odot (X_{\text{ctx}} - X) \|_2^2}{\sum M_{\text{bg}}} + \lambda_{\text{tgt}} \frac{\| M_{\text{tgt}} \odot X_{\text{ctx}} \|_2^2}{\sum M_{\text{tgt}}}
\label{eq:2}
\end{equation}
where $\odot$ denotes the Hadamard product and $\|\cdot\|_2^2$ represents the squared $L_2$ norm. The hyperparameters $\lambda_{\text{bg}}$ and $\lambda_{\text{tgt}}$ balance background reconstruction fidelity against target region suppression, and the denominators normalize each term by the effective area of its corresponding spatial mask.

The \textit{Concept Reconstruction Loss} ($\mathcal{L}_{\text{cpt}}$) ensures the Concept Branch bears sole responsibility for encoding target region semantics. Since the Context Branch is suppressed within $M_{\text{tgt}}$, the combined output of both branches is constrained to recover the original hidden state within this region:
\begin{equation}
\mathcal{L}_{\text{cpt}} = \lambda_{\text{cpt}} \frac{\| M_{\text{tgt}} \odot (X_{\text{ctx}} + X_{\text{cpt}} - X) \|_2^2}{\sum M_{\text{tgt}}}
\label{eq:3}
\end{equation}
where $\lambda_{\text{cpt}}$ weights the reconstruction fidelity within the target region. Together, $\mathcal{L}_{\text{ctx}}$ and $\mathcal{L}_{\text{cpt}}$ establish a complementary reconstruction protocol: the Context Branch covers the background while the Concept Branch covers the target region, and their union faithfully recovers the complete hidden state.

The \textit{Identity Leakage Penalty} ($\mathcal{L}_{\text{leak}}$) enforces semantic purity across the concept dictionary. We construct a binary penalty tensor $\mathbf{P}$ that equals $0$ for a concept's assigned kernels within $M_{\text{tgt}}$ and $1$ elsewhere:
\begin{equation}
\mathcal{L}_{\text{leak}} = \lambda_{\text{leak}} \frac{\| \mathbf{P} \odot f_{\text{cpt}} \|_1}{\sum \mathbf{P}}
\label{eq:4}
\end{equation}
where $\|\cdot\|_1$ denotes the $L_1$ norm, and $\lambda_{\text{leak}}$ controls the penalty strength. This strict spatial-semantic cross-regularization ensures that concept-specific feature kernels are strictly confined to their designated scope, eliminating false activations on background pixels or unassociated concept samples.

The \textit{Temporal Consistency Loss} ($\mathcal{L}_{\text{temp}}$) penalizes abrupt fluctuations in Context Branch activations across consecutive frames, since the scene canvas encoded by this branch must remain temporally stable to prevent flickering:
\begin{equation}
\mathcal{L}_{\text{temp}} = \lambda_{\text{temp}} \frac{\| f_{\text{ctx}}^{i} - f_{\text{ctx}}^{i-1} \|_2^2}{Z}
\label{eq:5}
\end{equation}
where $Z$ is a constant equal to the total feature volume, $i$ is the frame index. 

The \textit{Auxiliary Loss} ($\mathcal{L}_{\text{aux}}$) mitigates the dead latent problem~\cite{bussmann2024batchtopk} by encouraging inactive feature kernels across both branches to approximate the residual reconstruction error, ensuring high dictionary utilization. The overall training objective for PConvSAE is formulated as the sum of all five components:
\begin{equation}
\mathcal{L}_{\text{total}} = \mathcal{L}_{\text{ctx}} + \mathcal{L}_{\text{cpt}} + \mathcal{L}_{\text{leak}} + \mathcal{L}_{\text{temp}} + \mathcal{L}_{\text{aux}}
\label{eq:6}
\end{equation}

Through this joint optimization, PConvSAE decomposes entangled spatiotemporal representations into a structured, interpretable feature dictionary where concept-agnostic scene context and concept-specific semantics are cleanly separated into complementary branches. This decomposition provides the foundation for precise concept attribution and targeted erasure in the subsequent stages.

\subsection{Contrastive Attribution Mechanism}
\label{sec:attr}

The structured latent space established by PConvSAE provides a foundation for precise concept identification. The remaining challenge is to determine \emph{which} kernels within the Concept Branch are intrinsically bound to each target concept. Performing this identification dynamically at every inference step would introduce prohibitive computational overhead. We therefore decompose the process into a one-time offline attribution phase that locks concept-specific kernels prior to deployment, followed by an online mask-guided erasure stage in Sec.~\ref{sec:erase}.

The attribution phase aims to isolate, from the Concept Branch, the minimal subset of feature kernels that respond exclusively and consistently to a designated target concept $c_i$.  We first collect a set of positive samples generated from prompts containing $c_i$ and compute the spatial mean activation of each feature kernel strictly within the corresponding target region masks, yielding a per-kernel activation profile $\mu_{c_i} \in \mathbb{R}^{N_{\text{cpt}}}$. To ensure that the identified kernels are selective for $c_i$ rather than responsive to background structures or semantically adjacent but distinct concepts, we construct a hard-negative baseline $\mu_{\text{base}}$ that aggregates all competing activation sources.  Concretely, we compute $\mu_{\text{bg}}$, the global mean activation across pure background samples, together with $\mu_{c_j}$ for every non-target concept $c_j$ ($j \neq i$).  The hard-negative baseline is then defined as the element-wise maximum over all competing signals:
\begin{equation}
  \mu_{\text{base}} = \max\!\bigl(\mu_{\text{bg}},\;
        \max_{j \neq i}\,\mu_{c_j}\bigr)
  \label{eq:7}
\end{equation}
This formulation enforces a \emph{one-versus-max} exclusion principle: a kernel qualifies as concept-specific only if its response to $c_i$ exceeds both the background activation level and the strongest response elicited by any alternative concept.

We quantify the concept specificity of each kernel through a contrastive log-ratio attribution score $\mathcal{S}_{\text{attr}}$:
\begin{equation}
  \mathcal{S}_{\text{attr}} = \mu_{c_i} \odot
    \max\!\Bigl(0,\;\log\frac{\mu_{c_i}+\epsilon}
                               {\mu_{\text{base}}+\epsilon}\Bigr)
    \odot M_{\text{stable}}
  \label{eq:8}
\end{equation}
where $\epsilon$ is a small constant for numerical stability. The multiplicative weighting by $\mu_{c_i}$ ensures that kernels with higher absolute activation receive proportionally greater scores, favoring strongly responsive features over those that are selective yet weakly active. The binary mask $M_{\text{stable}}$ suppresses transient activations by retaining only temporally robust kernels.  To construct this mask, we compute the activation consistency of each kernel, defined as the fraction of positive samples in which the kernel successfully fires within the target region.  Kernels whose consistency falls below a predefined robustness threshold $\tau$ are discarded.

The top-$K$ kernels ranked by $\mathcal{S}_{\text{attr}}$ form an initial candidate set.  To preserve the architectural semantic isolation enforced during PConvSAE training, we intersect this candidate set with the pre-allocated index partition $\mathcal{I}_{\text{tgt}}$ assigned to concept $c_i$ in the channel partitioning scheme of the Concept Branch (Sec.~\ref{sec:pconvsae}).  The resulting set $\mathcal{K}_{\text{tgt}}$ constitutes the definitively locked concept kernels.  Because this offline attribution procedure is executed only once per target concept, it produces a compact kernel dictionary of high semantic purity while entirely eliminating the need for dynamic feature
search during inference.

\subsection{Concept Erasure with Dynamic Spatiotemporal Masks}
\label{sec:erase}

Diffusion features progressively transition from coarse global structure to fine-grained local detail across denoising timesteps.  This progressive evolution causes the spatial extent of a target concept to shift continuously throughout the reverse process.  Applying a static mask derived from a single timestep would therefore introduce severe pixel-level misalignments and boundary artifacts as features drift. To maintain tight spatial correspondence between the mask and the evolving concept representation, we propose a dynamic mask generation and intervention mechanism that adaptively updates the erasure region at each timestep~$t$.

\textbf{Dynamic Spatiotemporal Mask Generation.} Within a designated critical intervention interval, we pass the intermediate hidden states into the frozen PConvSAE at each timestep for online feature probing.  From the Context Branch, we extract the aggregated activation heatmap and compute its spatial complement to obtain a dynamic foreground mask $M_{\text{fg}}$.  Because the Context Branch is trained to encode concept-agnostic scene content (Eq.~\ref{eq:2}), its complement naturally highlights regions that fall outside the background canvas, providing a reliable foreground prior. Simultaneously, using the offline-locked kernel set $\mathcal{K}_{\text{tgt}}$, we extract the corresponding activation heatmap from the Concept Branch to produce the target concept mask $M_{\text{cpt}}$.  The final intervention mask $M_t$ is obtained as the Hadamard product of these two normalized components:
\begin{equation}
  M_{t} = \mathrm{Norm}(M_{\text{fg}}) \odot
           \mathrm{Norm}(M_{\text{cpt}})
  \label{eq:9}
\end{equation}
where $\mathrm{Norm}(\cdot)$ denotes min-max normalization. A dynamic threshold is subsequently applied to yield a binarized mask.  The intersection of $M_{\text{fg}}$ and $M_{\text{cpt}}$ provides complementary spatial constraints: $M_{\text{fg}}$ confines the mask to foreground regions, preventing spurious suppression of background content, while $M_{\text{cpt}}$ further restricts it to locations where the target concept is semantically active.  Because this mask is recomputed at every timestep, it adaptively tracks the spatial trajectory of the target concept across both frames and denoising stages.

\textbf{Spatially-Modulated Classifier-Free Guidance.} Equipped with the dynamic mask $M_t$, we integrate concept erasure directly into the classifier-free guidance (CFG) mechanism. Standard CFG applies guidance uniformly across all spatial positions, which inadvertently perturbs global illumination, background structure, and temporal dynamics when repurposed for concept suppression.  We instead propose \emph{spatially-modulated CFG}, which restricts negative guidance exclusively to the regions delineated by the mask:
\begin{equation}
  \hat{\epsilon}_{\theta}(x_t) =
    \epsilon_{\theta}(x_t, c_{\text{pos}})
    - s \cdot \bigl(\epsilon_{\theta}(x_t, c_{\text{pos}})
                   - \epsilon_{\theta}(x_t, c_{\text{neg}})\bigr)
      \odot M_t
  \label{eq:10}
\end{equation}
Here $\epsilon_{\theta}(x_t, c_{\text{pos}})$ and $\epsilon_{\theta}(x_t, c_{\text{neg}})$ denote the noise predictions conditioned on the positive prompt containing the target concept and the negative prompt providing a safe semantic alternative, respectively, and $s$ controls the erasure guidance scale.  Within regions where $M_t \approx 1$, the guidance term steers the denoising trajectory away from the target semantics, imposing localized concept suppression.  In regions where $M_t \approx 0$, the original positive-conditioned prediction is preserved without modification, maintaining the integrity of background content. 

\section{Experiments}
We evaluate \textit{EraseSAE} across two concept erasure tasks, nudity erasure and celebrity identity erasure, on two DiT-based T2V models: CogVideoX-5B~\cite{yang2025cogvideox} and HunyuanVideo~\cite{kong2024hunyuanvideo}. To further validate the generality of our framework beyond the video domain, we conduct additional nudity erasure experiments on the widely adopted T2I model Flux.1 [dev]~\cite{flux2024}. Unless stated otherwise, PConvSAE is trained with the Adam optimizer~\cite{kingma2014adam} at an initial learning rate of $1\times10^{-4}$ for 30 epochs on 8 NVIDIA A100 GPUs. More implementation details are provided in the supplementary material.

\begin{table*}[!t]
\centering
\caption{Quantitative comparison of nudity erasure methods on CogVideoX and HunyuanVideo. Best and second best results are \textbf{bolded} and \underline{underlined}.}
\renewcommand{\arraystretch}{1.2} 
\adjustbox{max width=0.98\textwidth}{
\begin{tabular}{l *{14}{c}|*{2}{c}}
\toprule 
\multirow{3}{*}{\textbf{Method}} & 
\multicolumn{2}{c}{\multirow{2}{*}{\begin{tabular}[c]{@{}c@{}}\textbf{Nudity} \\ \textbf{Rate (Gen)} ($\downarrow$)\end{tabular}}} & 
\multicolumn{2}{c}{\multirow{2}{*}{\begin{tabular}[c]{@{}c@{}}\textbf{Object} \\ \textbf{Class} ($\uparrow$)\end{tabular}}} & 
\multicolumn{2}{c}{\multirow{2}{*}{\begin{tabular}[c]{@{}c@{}}\textbf{Subject} \\ \textbf{Consistency} ($\uparrow$)\end{tabular}}} & 
\multicolumn{2}{c}{\multirow{2}{*}{\textbf{SSIM} ($\uparrow$)}} & 
\multicolumn{6}{c}{\textbf{Nudity Rate (Ring-A-Bell)} ($\downarrow$)} &
\multicolumn{2}{|c}{\multirow{2}{*}{\begin{tabular}[c]{@{}c@{}}\textbf{Inference Time} \\ \textbf{(s/frame)} \end{tabular}}} \\
\cmidrule(lr){10-15}
& \multicolumn{2}{c}{} & \multicolumn{2}{c}{} & \multicolumn{2}{c}{} & \multicolumn{2}{c}{} & 
\multicolumn{2}{c}{K16} & \multicolumn{2}{c}{K38} & \multicolumn{2}{c}{K77} & \multicolumn{2}{|c}{}\\ 
\cmidrule(lr){2-3} \cmidrule(lr){4-5} \cmidrule(lr){6-7} \cmidrule(lr){8-9} \cmidrule(lr){10-11} \cmidrule(lr){12-13} \cmidrule(lr){14-15} \cmidrule(lr){16-17}
& CogX & Hunyuan & CogX & Hunyuan & CogX & Hunyuan & CogX & Hunyuan & CogX & Hunyuan & CogX & Hunyuan & CogX & Hunyuan & CogX & Hunyuan\\ 
\midrule 
Original & 28.25 & 68.13 & \textbf{79.91} & \underline{83.88} & 95.12 & \textbf{96.12} & - & - & 14.47 & 21.97 & 20.26 & 30.53 & 29.34 & 23.55 & \textbf{3.61} & \textbf{3.55}\\
Neg Prompt & 27.75 & 57.25 & 75.68 & \textbf{87.59} & \underline{95.87} & 95.48 & \underline{47.02} & 47.06 & 10.79 & 22.37 & 34.34 & 29.34 & 38.03 & 31.97 & 3.66 & 6.95\\ 
SAFREE~\cite{yoon2025safree} & 5.75 & 46.38 & 45.19 & 71.47 & 94.49 & 95.33 & 42.39 & 41.78 & 15.13 & 36.05 & 16.58 & 31.71 & 15.66 & 19.74 & 3.72 & \underline{3.59}\\
VideoEraser~\cite{xu2025videoeraser} & 18.88 & - & 73.86 & - & \textbf{96.47} & - & 44.36 & - & 7.11 & - & 18.03 & - & 23.68 & - & 4.03 & -\\
T2VUnlearning~\cite{ye2025t2vunlearning} & \underline{2.88} & \underline{10.89} & 51.98 & 77.33 & 93.55 & 95.96 & 42.39 & \underline{60.67} & \underline{5.30} & \underline{5.13} & \underline{7.60} & \underline{7.82} & \underline{8.65} & \underline{9.95} & 3.67 & 3.60\\
\rowcolor{gray!15} \textbf{Ours} & \textbf{2.62} & \textbf{7.13} & \underline{77.80} & 80.61 & 94.30 & \underline{96.04} & \textbf{74.99} & \textbf{65.69} & \textbf{2.63} & \textbf{4.47} & \textbf{5.26} & \textbf{6.84} & \textbf{4.74} & \textbf{5.13} & \underline{3.66} & 6.97\\
\bottomrule 
\end{tabular}}
\label{tab:nudity}
\end{table*}

\subsection{Nudity Erasure}
\label{sec:ne}

\textbf{Experimental Settings.} Following~\cite{ye2025t2vunlearning}, we construct a dataset of 500 nudity-related prompts, allocating 400 for training and 100 for testing, denoted as \textbf{Gen}. To evaluate defensive robustness against adversarial prompt manipulation, we additionally employ the adversarial prompt benchmark \textbf{Ring-A-Bell}~\cite{hsu2024ring}. For T2I, we conduct on Flux.1 [dev] using the \textbf{I2P}~\cite{schramowski2023safe} dataset. All output videos are standardized to 32 frames at 8 fps. For quantitative evaluation, we employ \textbf{NudeNet}~\cite{bedapudi2019nudenet} to perform frame-by-frame detection and compute the target nudity exposure rate. To assess non-destructive fidelity, we adopt the Structural Similarity (\textbf{SSIM})~\cite{wang2004image} to quantify pixel-level consistency against unaltered reference videos. We further incorporate the \textbf{VBench}~\cite{huang2024vbench} Object Class and Subject Consistency metrics to verify the coherent preservation of non-target semantics.

\textbf{Quantitative Results.} Tab.~\ref{tab:nudity} presents comprehensive comparisons with state-of-the-art methods on the video nudity erasure task. Our method achieves the lowest nudity exposure rates on both CogVideoX-5B and HunyuanVideo, reducing the Gen-set exposure to 2.62 and 7.13, respectively. This consistent superiority over the strongest baseline, T2VUnlearning~\cite{ye2025t2vunlearning}, validates the effectiveness of operating in the monosemantic feature space: because PConvSAE decomposes entangled activations into concept-specific sparse features, the resulting concept kernels capture the target semantics with high purity, enabling thorough suppression that weight-editing approaches struggle to achieve.

The advantage of \textit{EraseSAE} is particularly pronounced under adversarial evaluation. On the Ring-A-Bell benchmark across all three difficulty levels (K16, K38, K77), our method maintains consistently low detection rates. We attribute this robustness to the contrastive attribution mechanism, which locks concept kernels through one-versus-max exclusion against hard negatives and thereby captures intrinsic concept semantics rather than surface-level prompt patterns. Consequently, adversarial prompt reformulations fail to circumvent the erasure because the underlying monosemantic features remain suppressed regardless of input phrasing. As shown in Tab.~\ref{tab:flux-i2p}, \textit{EraseSAE} further extends its efficacy to the T2I model Flux.1 [dev], achieving the lowest total nudity count on the challenging I2P benchmark and demonstrating the architectural generality.

Crucially, this strong erasure performance incurs minimal collateral damage. Our method achieves SSIM of 74.99 and 65.69 on CogVideoX-5B and HunyuanVideo, surpassing the next best method by 27.97 and 5.02 points. This gap reflects the spatiotemporal locality afforded by the dynamic mask mechanism, which confines interventions to active concept regions and preserves the pixel-level structure that global editing inevitably degrades. The VBench Object Class and Subject Consistency scores remain competitive with the unmodified models, confirming that PConvSAE's partitioned architecture disentangles concept-specific features from scene context. \textit{EraseSAE} further introduces negligible inference overhead. On CogVideoX-5B its latency reaches 3.66 s/frame against 3.61 for the original model, while the marginal cost on HunyuanVideo originates solely from the additional negative-conditioned forward pass and stays on par with standard negative prompting. Because concept attribution is conducted entirely offline, no per-step feature search is required during inference.

\begin{figure}[t]
\centering
\captionsetup{font=small}
\includegraphics[width=0.95\textwidth]{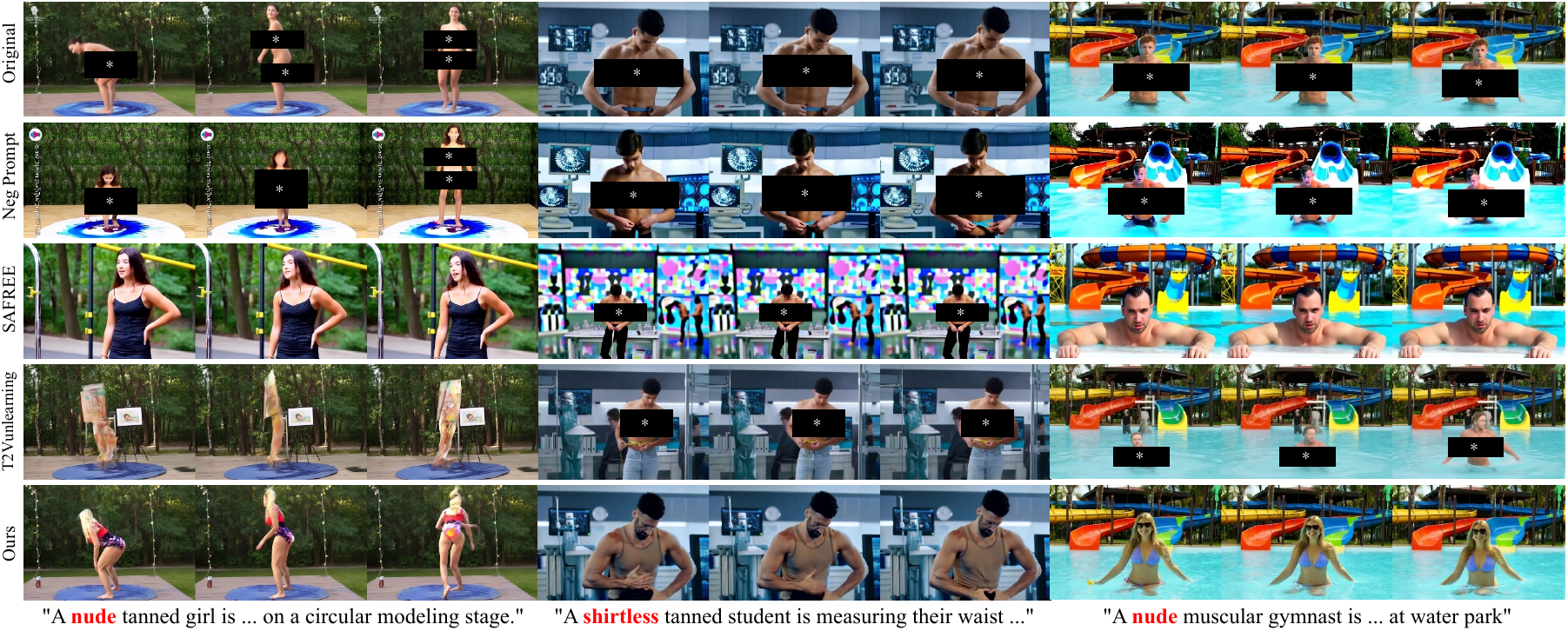}
\caption{Qualitative comparison of nudity erasure methods on HunyuanVideo.}
\label{fig:q_nud}
\end{figure}

\textbf{Qualitative Results.}
Fig.~\ref{fig:q_nud} provides visual comparisons between \textit{EraseSAE} and prior methods. Training-free approaches such as Neg Prompt and SAFREE~\cite{yoon2025safree} fail to completely suppress sensitive content. T2VUnlearning~\cite{ye2025t2vunlearning} achieves stronger erasure through weight editing, yet introduces visible artifacts in background regions and distorts non-target objects due to global parameter modifications. In contrast, \textit{EraseSAE} cleanly removes the target concept while faithfully preserving unrelated visual content, demonstrating the practical benefit of spatially-modulated erasure guided by monosemantic feature maps.

\subsection{Celebrity Erasure}
\label{sec:ce}

\textbf{Experimental Settings.}
To evaluate identity protection capabilities, we conduct erasure experiments targeting five prominent public figures: \textit{Donald Trump, Barack Obama, Elon Musk, Queen Elizabeth, and Taylor Swift}. Each category comprises 400 prompts, equally split between training and testing. Our evaluation encompasses three complementary assessments. The primary assessment evaluates standard identity erasure efficacy by measuring the residual detection accuracy of the target individual. The secondary assessment quantifies cross-identity interference: we apply the erasure intervention for a single target celebrity while concurrently generating videos of the remaining four celebrities, and compute the average detection accuracy of these unedited identities to measure their retention rate. The tertiary assessment employs SSIM to quantify the preservation of non-target backgrounds and general context. Target identity presence is measured using the Giphy celebrity detection algorithm~\cite{xu2025videoeraser}.

\begin{table*}[!t]
\centering
\caption{Quantitative results of celebrity erasure on CogVideoX and HunyuanVideo.}
\renewcommand{\arraystretch}{1.1} 
\adjustbox{max width=0.95\textwidth}{
\begin{tabular}{ll *{12}{c} *{1}{|c}*{1}{c}}
\toprule 
\multirow{2}{*}{\textbf{Method}} & \multirow{2}{*}{\textbf{Metric}} & 
\multicolumn{2}{c}{\textbf{Donald Trump}} & \multicolumn{2}{c}{\textbf{Barack Obama}} & \multicolumn{2}{c}{\textbf{Elon Musk}} & 
\multicolumn{2}{c}{\textbf{Queen Elizabeth}} & \multicolumn{2}{c}{\textbf{Taylor Swift}} & \multicolumn{2}{c}{\textbf{Average}} & \multicolumn{2}{|c}{\textbf{Preserve $\uparrow$}}  \\ 
\cmidrule(lr){3-4} \cmidrule(lr){5-6} \cmidrule(lr){7-8} \cmidrule(lr){9-10} \cmidrule(lr){11-12} \cmidrule(lr){13-14} \cmidrule(lr){15-16}
& & CogX & Hunyuan & CogX & Hunyuan & CogX & Hunyuan  & CogX & Hunyuan  & CogX & Hunyuan  &  CogX & Hunyuan &  CogX & Hunyuan \\ 
\midrule 
\multirow{1}{*}{Original} 
& Erase ($\downarrow$) & 77.50 & 87.50 & 72.50 & 75.00 & 49.50 & 53.50 & 68.50 & 73.00 & 49.00 & 46.50 & 63.40 & 67.10 & - & - \\
\midrule
\multirow{2}{*}{Neg Prompt} 
& Erase ($\downarrow$) & 63.50 & 59.00 & 69.00 & 53.00 & 24.50 & 32.50 & 29.50 & 50.00 & 26.00 & 41.50 & 42.50 & 47.20 & \multirow{2}{*}{\textbf{59.80}} & \multirow{2}{*}{60.20}\\
& SSIM ($\uparrow$)   &  31.30   &  35.31  &  32.58   &  35.31   &  33.57   &  43.76   &  27.44   &  33.02   &  32.58   &  40.13   &  31.49   &  37.51   \\
\midrule
\multirow{2}{*}{SAFREE~\cite{yoon2025safree}} 
& Erase ($\downarrow$) & 62.50 & \textbf{9.50} & 67.00 & 13.50 & 17.50 & 53.00 & 21.50 & 68.00 & 19.50 & 50.50 & 37.60 & 47.20 & \multirow{2}{*}{48.40} & \multirow{2}{*}{47.00}\\
& SSIM ($\uparrow$)   &  30.74   &  31.40   &  30.37   &  30.17   &  34.08   &  39.54   &  27.26   &  32.36   &  31.87   &  32.26  &  30.86   &  33.14   \\
\midrule
\multirow{2}{*}{VideoEraser~\cite{xu2025videoeraser}} 
& Erase ($\downarrow$) & \textbf{3.00} & - & 20.00 & - & 19.00 & - & 41.00 & - & 42.00 & - & 25.00 & - & \multirow{2}{*}{52.10} & \multirow{2}{*}{-}  \\
& SSIM ($\uparrow$)   &  29.32   &  -   &  28.11   &  -   &  30.73   &  -   &  22.85   &  -   &  25.73   &  -   &  27.35   &  -   \\
\midrule
\rowcolor{gray!15}  
& Erase ($\downarrow$) & 12.00 & 26.50 & \textbf{1.50} & \textbf{11.50} & \textbf{6.50} & \textbf{11.50} & \textbf{16.50} & \textbf{26.00} & \textbf{3.50} & \textbf{20.50} & \textbf{8.00} & \textbf{19.20} & & \\
\rowcolor{gray!15} \multirow{-2}{*}{\textbf{Ours}}
& SSIM ($\uparrow$)   &  \textbf{63.12}   &  \textbf{79.53}   & \textbf{61.5}   &  \textbf{62.12}   &  \textbf{66.35}   &  \textbf{85.54}   &  \textbf{61.14}   &  \textbf{75.89}   &  \textbf{64.81}   &  \textbf{84.27}   &  \textbf{63.38}  &  \textbf{77.47}  & \multirow{-2}{*}{58.60} & \multirow{-2}{*}{\textbf{61.50}}\\
\bottomrule 
\end{tabular}}
\label{tab:celebrity}
\end{table*}

\textbf{Quantitative Results.}
Tab.~\ref{tab:celebrity} presents the comprehensive evaluation of celebrity identity erasure. Our framework reduces the average target detection accuracy to 8.00 and 19.20 on CogVideoX-5B and HunyuanVideo, respectively, demonstrating effective identity suppression across architectures and diverse facial characteristics. While the target identity is thoroughly suppressed, \textit{EraseSAE} maintains average preservation accuracies of 58.60 and 61.50 for non-target identities on the two models, closely matching or exceeding the preservation scores of all baselines. This outcome validates the strict semantic exclusivity enforced by the one-versus-max attribution scoring (Eq.~\ref{eq:8}) and the channel partitioning scheme within the Concept Branch: the feature kernels locked for one identity reside in a mutually exclusive subspace from those encoding other identities, preventing cross-concept interference during erasure. In contrast, VideoEraser~\cite{xu2025videoeraser}, despite achieving the lowest erasure rate for Donald Trump (3.00), attains only 52.10 preservation accuracy, indicating that its text-embedding manipulation inadvertently corrupts shared facial representations. The SSIM scores further underscore the non-destructive nature of our approach. 

\begin{figure}[t]
    \centering
    \begin{minipage}{0.35\textwidth}
        \centering
        \includegraphics[width=0.92\textwidth]{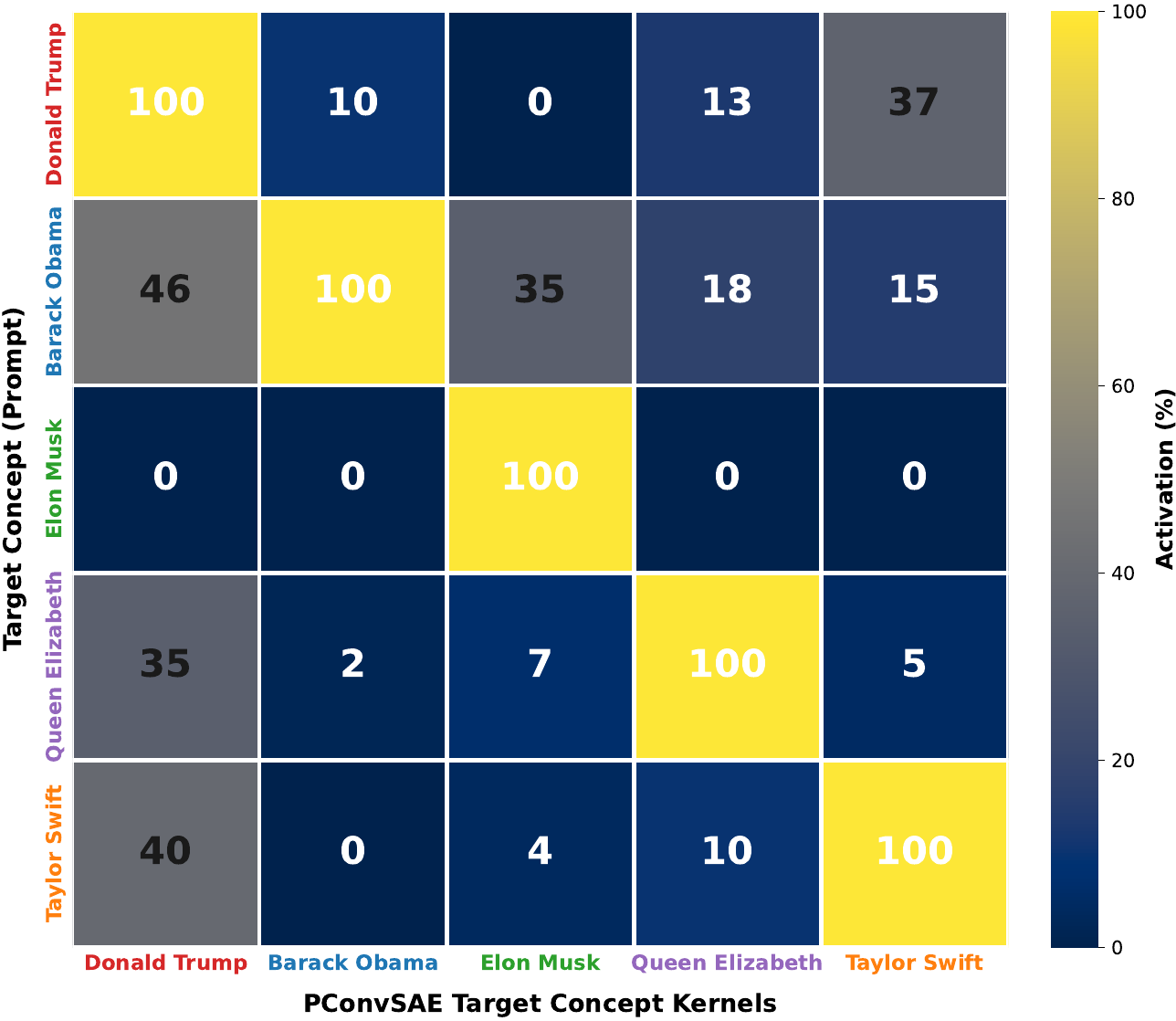} 
        \caption{Cross-identity interference heatmap of 5 celebrities.}
        \label{fig:multi_concept}
    \end{minipage}
    \hfill
    \begin{minipage}{0.62\textwidth}
        \centering
        \includegraphics[width=0.98\textwidth]{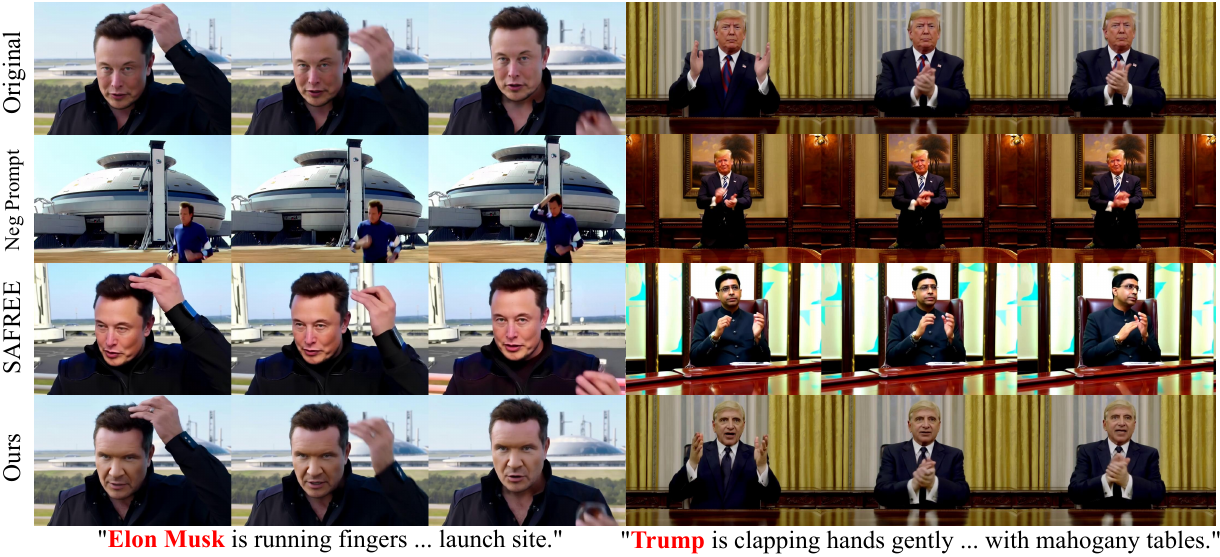} 
        \caption{Qualitative comparison of celebrity erasure methods on HunyuanVideo.}
        \label{fig:q_celeb}
    \end{minipage}
\end{figure}

To validate the monosemantic property of the co-learned concept kernels, we present a cross-identity interference heatmap in Fig.~\ref{fig:multi_concept}. For each concept's dedicated kernel set, we compute the raw activation magnitude in response to prompts targeting each of the five celebrity identities and normalize the values to a percentage scale. The heatmap exhibits strong diagonal dominance, confirming that each kernel set responds selectively and exclusively to its designated target identity, with minimal cross-activation to other identities. Beyond cross-identity interference, we directly quantify the semantic purity of locked kernels~\cite{templeton2024scaling}. We define the \textit{purity} of class $c$ as $P_c = \frac{1}{|\mathcal{K}_{\text{tgt}}^{(c)}|}\sum_{k \in \mathcal{K}_{\text{tgt}}^{(c)}} \mathrm{Act}_{c}(k) / \sum_{c'=1}^{C}\mathrm{Act}_{c'}(k)$, where $\mathrm{Act}_{c}(k)$ is the spatiotemporal mean of the post-Top-K activation of kernel $k$ over 100 samples from class $c$. Averaged over the 5 identities, PConvSAE attains $\bar{P}=0.76$, more than twice the $0.37$ of a linear SAE, confirming high monosemanticity.

\textbf{Qualitative Results.} Qualitative results in Fig.~\ref{fig:q_celeb} corroborate the quantitative findings. Neg Prompt and SAFREE exhibit inconsistent erasure, with recognizable facial features of the target identity persisting across multiple frames. In contrast, \textit{EraseSAE} completely suppresses the target celebrity's distinguishing features while faithfully preserving the fidelity of complex background.

\subsection{Ablation Studies}
\label{sec:ab}

All ablation studies are conducted on HunyuanVideo for the nudity erasure task.

\textbf{Loss Components.} Tab.~\ref{tab:ablation_losses} reports a cumulative ablation of the PConvSAE objectives. With only $\mathcal{L}_{\text{ctx}}$, the Concept Branch lacks explicit guidance for isolating target semantics, yielding a high nudity rate of 24.5. Sequentially introducing $\mathcal{L}_{\text{cpt}}$ and $\mathcal{L}_{\text{leak}}$ strengthens semantic separation and concept purity, reducing the rate to 8.09. Adding $\mathcal{L}_{\text{temp}}$ further raises SSIM by 18.32 points to 65.69, confirming its role in stabilizing temporal coherence.

\textbf{Architectural Design.} Tab.~\ref{tab:ablation_architectures} evaluates three variants: (1) \textit{Linear SAE}~\cite{cywinski2025saeuron}, the traditional linear sparse autoencoder extended to T2V activations; (2) \textit{PConvSAE (single-branch)}, our convolutional encoder-decoder with a single undifferentiated branch; and (3) \textit{PConvSAE (dual-branch)}, the full architecture with separate Context and Concept Branches. The Linear SAE flattens spatiotemporal activations into one-dimensional vectors, destroying local structure needed for precise localization and yielding the weakest results (12.76 nudity rate, 56.97 SSIM). The single-branch variant preserves spatial topology and improves both metrics, yet residual concept-context entanglement persists without explicit partitioning. The dual-branch PConvSAE attains the best performance on all metrics, demonstrating that separating concept-specific and context-agnostic representations is essential for both thorough erasure and high-fidelity preservation.

\textbf{Inference Strategy.} Tab.~\ref{tab:ablation_erasure} compares three inference-time interventions. SAE-Sub directly subtracts target feature activations and suppresses concepts insufficiently (51.30 nudity rate). SAE-Mask zeros out all activations within the masked regions across layers; this aggressive operation lowers the nudity rate to 6.56 but corrupts the scene context encoded therein, degrading SSIM to 53.51. Our spatially-modulated classifier-free guidance (SM-CFG) attains the optimal trade-off between erasure thoroughness and structural preservation.

\begin{table}[t]
    \centering
    \renewcommand{\arraystretch}{0.9}
    \begin{minipage}[t]{0.5\textwidth}
        \centering
        \caption{Quantity of explicit content ($\downarrow$) with Flux.1 on I2P dataset.}
        \label{tab:flux-i2p}
        \begin{adjustbox}{width=0.88\linewidth}
            \setlength{\tabcolsep}{4pt}
            \begin{tabular}{lcccc}
                \toprule
                Method        & Common & Female & Male & Total \\
                \midrule
                Original      & 406 & 161 & 38 & 605 \\
                UCE~\cite{gandikota2024unified} & 122 & 39  & 12 & 173 \\
                MACE~\cite{lu2024mace}          & 173 & 55  & 28 & 256 \\
                EAP~\cite{bui2024erasing}       & 287 & 86  & 13 & 386 \\
                EraseAnything~\cite{gao2025eraseanything} & 129 & 48 & \textbf{22} & 199 \\
                \rowcolor{gray!15}
                \textbf{Ours} & \textbf{102} & \textbf{31} & 27 & \textbf{160} \\
                \bottomrule
            \end{tabular}
        \end{adjustbox}
    \end{minipage}
    \hfill
    \begin{minipage}[t]{0.45\textwidth}
        \centering
        \caption{Ablation study on the impact of each loss function design.}
        \label{tab:ablation_losses}
        \begin{adjustbox}{width=0.9\linewidth}
            \setlength{\tabcolsep}{4pt}
            \begin{tabular}{cccc|cc}
                \toprule
                $\mathcal{L}_\text{ctx}$ & $\mathcal{L}_\text{cpt}$ & $\mathcal{L}_\text{leak}$ & $\mathcal{L}_\text{temp}$ & Nude Rate $\downarrow$ & SSIM $\uparrow$ \\
                \midrule
                \checkmark &            &            &            & 24.5  & 40.12 \\
                \checkmark & \checkmark &            &            & 10.21 & 43.33 \\
                \checkmark & \checkmark & \checkmark &            & 8.09  & 47.37 \\
                \checkmark & \checkmark & \checkmark & \checkmark & \textbf{7.13} & \textbf{65.69} \\
                \bottomrule
            \end{tabular}
        \end{adjustbox}
    \end{minipage}
\end{table}

\begin{table}[t]
    \centering
    \renewcommand{\arraystretch}{0.9}
    \begin{minipage}[t]{0.5\textwidth}
        \centering
        \caption{Ablation study on different SAE architectural variants.}
        \label{tab:ablation_architectures}
        \begin{adjustbox}{width=0.88\linewidth}
            \setlength{\tabcolsep}{4pt}
            \begin{tabular}{lcc}
                \toprule
                Method & Nude Rate $\downarrow$ & SSIM $\uparrow$ \\
                \midrule
                Linear SAE~\cite{cywinski2025saeuron} & 12.76 & 56.97 \\
                PConvSAE (single-branch) & 9.06 & 61.37 \\
                \rowcolor{gray!25} PConvSAE (dual-branch) & \textbf{7.13} & \textbf{65.69} \\
                \bottomrule
            \end{tabular}
        \end{adjustbox}
    \end{minipage}
    \hfill
    \begin{minipage}[t]{0.45\textwidth}
        \centering
        \caption{Ablation study on different inference strategies.}
        \label{tab:ablation_erasure}
        \begin{adjustbox}{width=0.68\linewidth}
            \setlength{\tabcolsep}{4pt}
            \begin{tabular}{lcc}
                \toprule
                Method & Nude Rate $\downarrow$ & SSIM $\uparrow$ \\
                \midrule
                SAE-Sub  & 51.30 & 62.31 \\
                SAE-Mask & \textbf{6.56} & 53.51 \\
                \rowcolor{gray!25}
                SM-CFG   & 7.13 & \textbf{65.69} \\
                \bottomrule
            \end{tabular}
        \end{adjustbox}
    \end{minipage}
\end{table}

\section{Conclusion}
In this paper, we introduce \textit{EraseSAE}, a pioneering framework for surgical concept erasure in T2V diffusion models. By shifting the intervention paradigm from entangled polysemantic neurons to disentangled monosemantic features, we resolve the granularity mismatch plaguing prior approaches. Our proposed PConvSAE effectively decomposes dense spatiotemporal activations into interpretable units while preserving essential structural coherence. Coupled with a novel contrastive attribution mechanism, our \textit{decompose-attribute-erase} pipeline precisely isolates and suppresses target concepts exclusively within their active regions. Extensive evaluations demonstrate that \textit{EraseSAE} achieves state-of-the-art erasure efficacy with minimal degradation to overall generation quality. 

\section*{Acknowledgments}
This work was supported by the Key Science \& Technology Project of Anhui Province No. 202523o09050002, National Natural Science Foundation of China No. 62472396, Anhui Provincial Natural Science Foundation No. 2508085QF212, and Fundamental Research Funds for the Central Universities No. WK2102026003.

%
%
\bibliographystyle{splncs04}
\bibliography{main}
\end{document}